\documentclass[10pt,twocolumn,letterpaper]{article}

\usepackage{wacv}
\usepackage{times}
\usepackage{epsfig}
\usepackage{graphicx}
\usepackage{amsmath}
\usepackage{amssymb}
\usepackage{booktabs}

\usepackage{graphicx}
\usepackage{booktabs}
\usepackage{multirow}
\usepackage{float}
\usepackage{booktabs}
\usepackage{makecell}
\usepackage{subfig}
\usepackage{xspace}
\usepackage{animate}

\newcounter{FPS}
\ifwacvfinal
\usepackage[breaklinks=true,bookmarks=false]{hyperref}
\else
\usepackage[pagebackref=true,breaklinks=true,colorlinks,bookmarks=false]{hyperref}
\fi

\makeatletter
\DeclareRobustCommand\onedot{\futurelet\@let@token\@onedot}
\def\@onedot{\ifx\@let@token.\else.\null\fi\xspace}

\def\eg{\emph{e.g}\onedot} 
\def\ie{\emph{i.e}\onedot}

\def\etal{\emph{et al}\onedot}
\makeatother

\usepackage[capitalize]{cleveref}
\crefname{section}{Sec.}{Secs.}
\Crefname{section}{Section}{Sections}
\Crefname{table}{Table}{Tables}
\crefname{table}{Tab.}{Tabs.}

\newcommand{\stereo}[2]{
  \ensuremath{\mathbf{#1}^{#2}_\text{S}}
}
\newcommand{\motion}[2]{
  \ensuremath{\mathbf{#1}^{#2}_\text{M}}
}
\newcommand{\fusion}[2]{
  \ensuremath{\mathbf{#1}^{#2}_\text{F}}
}
\newcommand{\xstereo}[1]{\ensuremath{#1_\text{S}}}
\newcommand{\xmotion}[1]{\ensuremath{#1_\text{M}}}
\newcommand{\xfusion}[1]{\ensuremath{#1_\text{F}}}

\newcommand{\N}{\mathcal{N}}

\newcommand{\trans}{\ensuremath{{\textbf{\textit{T}}}}}
\newcommand{\wreset}{\ensuremath{\mathbf{w}_\text{reset}}}
\newcommand{\wfusion}{\ensuremath{\mathbf{w}_\text{fusion}}}

\newcommand{\ours}{CODD}
\newcommand{\epe}{\ensuremath{\text{EPE}}}
\newcommand{\teper}{\ensuremath{\text{TEPE}_\text{r}}}
\newcommand{\tepe}{\ensuremath{\text{TEPE}}}
\newcommand{\fepenull}{\ensuremath{\text{FEPE}}}
\newcommand{\fepe}[1]{\ensuremath{\text{FEPE}^{\text{#1}}}}
\newcommand{\dmetric}[1]{\ensuremath{\delta_{\text{#1}}}}
\newcommand{\dmetricsup}[2]{\ensuremath{\delta_{\text{#1}}^\text{#2}}}

\newcommand{\aref}[1]{\hyperref[#1]{Appx~\ref*{#1}}}

\newcommand{\mli}[1]{\textcolor{black}{#1}}

\def\wacvPaperID{xxx} 

\wacvapplicationstrack 

\wacvfinalcopy 

\usepackage[accsupp]{axessibility}

\begin{document}

\title{Temporally Consistent Online Depth Estimation in Dynamic Scenes}

\author{%
Zhaoshuo Li$^{1}$ \quad Wei Ye$^{2}$ \quad Dilin Wang$^{2}$ \quad Francis X. Creighton$^{1}$ \\
Russell H. Taylor$^{1}$ \quad Ganesh Venkatesh$^{2}$ \quad Mathias Unberath$^{1}$
\vspace{0.05in}\\
$^{1}$ Johns Hopkins University \hspace{.2in}
$^{2}$ Reality Labs, Meta Inc.
}

\maketitle
\thispagestyle{empty}

\begin{abstract}
    Temporally consistent depth estimation is crucial for online applications such as augmented reality. While stereo depth estimation has received substantial attention as a promising way to generate 3D information, there is relatively little work focused on maintaining temporal stability. Indeed, based on our analysis, current techniques still suffer from poor temporal consistency. Stabilizing depth temporally in dynamic scenes is challenging due to concurrent object and camera motion. In an online setting, this process is further aggravated because only past frames are available.
We present a framework named Consistent Online Dynamic Depth (\ours{}) to produce temporally consistent depth estimates in dynamic scenes in an online setting. \ours{} augments per-frame stereo networks with novel motion and fusion networks. The motion network accounts for dynamics by predicting a per-pixel SE3 transformation and aligning the observations. The fusion network improves temporal depth consistency by aggregating the current and past estimates.
We conduct extensive experiments and demonstrate quantitatively and qualitatively that \ours{} outperforms competing methods in terms of temporal consistency and performs on par in terms of per-frame accuracy. 
Project page:
{\selectfont \url{https://mli0603.github.io/codd}}
\end{abstract}

\section{Introduction}
\label{sec:intro}

\begin{figure*}[t]
    \centering
    \animategraphics[loop,controls=play,width=\linewidth]{\theFPS}{img/animation/frame_}{25}{61}
    \caption{Temporal consistency comparison of per-frame stereo depth estimation solutions and our proposed \ours{} framework. The point cloud visualization is generated given the depth estimates. \textbf{This animated figure is best viewed in Adobe Reader (click the button to play).} More visualizations can be found in the video supplement.}
    \label{fig:frame_frame_jitter}
\end{figure*}

For online applications such as augmented reality, estimating consistent depth across video sequences is important, as temporal noise in depth estimation may corrupt visual quality and interfere with downstream processing such as surface extraction. One way to acquire metric depth (\ie without scale ambiguity) is to use calibrated stereo images. Recent developments in stereo depth estimation have been focusing on improving disparity accuracy on a per-frame basis \cite{hirschmuller2007stereo,chang2018pyramid,guo2019group,tankovich2021hitnet,li2021revisiting}. However, none of these approaches considers temporal information nor attempts to maintain temporal consistency. We examine the temporal stability of different per-frame networks and find that current solutions suffer from poor temporal consistency. We quantify such temporal inconsistency in predictions in \autoref{sec:experimental_setup} and provide qualitative visualization of resulting artifacts in the video supplement to further illustrate the case. 

We posit that stabilizing depth estimation temporally requires reasoning between current and previous frames, \ie establishing cross-frame correspondences and correlating the predicted values. In the simplest case where the scene is entirely static and camera poses are known \cite{liu2019neural,long2021multi}, camera motion can be corrected by a single SE3 transformation. Considering geometric constraints of multiple camera views \cite{andrew2001multiple}, the aligned cameras have the same viewpoint onto the static scene and therefore, depth values are expected to be the same, which allows pixel-wise aggregation of depth estimates from different times for consistency. 

However, in a \textit{dynamic} environment with moving and deforming objects, multi-view constraints do not hold. Even if cross-frame correspondences are established, independent depth estimates for corresponding points cannot simply be fused. This is because depth is not translation invariant and thus, fusion requires aligning depth predictions into a common coordinate frame. Therefore, prior works \cite{luo2020consistent,kopf2021robust} explicitly remove moving objects and only stabilize the static background to comply with the constraint. Given additional 3D motion, \eg estimated by scene flow, the depth of both moving and static objects can be aligned, which then enables temporal consistency processing \cite{zhang2021consistent}. However, as do the previously mentioned approaches, Zhang \etal \cite{zhang2021consistent} use information from all video frames and optimize network parameters at application time, limiting itself to \textit{offline} use. 

In an \textit{online} setting, prior works \cite{tananaev2018temporally,zhang2019exploiting,eom2019temporally} incorporate temporal information by appending a recurrent network to the depth estimation modules. However, these recurrent networks do not provide explicit reasoning between frames. Moreover, these approaches \cite{tananaev2018temporally,zhang2019exploiting,eom2019temporally} consider depth estimation from single images, not stereo pairs. Due to the scale ambiguity of monocular depth estimation, prior works mainly focus on producing estimates with consistent scale  \cite{kopf2021robust,zhang2021consistent} instead of reducing the inter-frame jitter that arises in metric depth estimation.

Traditionally, to encourage temporal consistency in \textit{metric depth}, prior techniques have relied on hand-crafted probability weights \cite{kalman1960new,matyunin2011temporal,tsekourakis2019measuring}. One example is the Kalman filter \cite{kalman1960new}, which combines previous and current predictions based on associated uncertainties. However, it assumes Gaussian measurement error, which often fails in scenarios with occlusion and de-occlusion between frames.

We present a framework named Consistent Online Dynamic Depth (\ours{}) that produces temporally consistent depth predictions. To account for inter-frame motion, we integrate a motion network that predicts a per-pixel SE3 transformation that aligns previous estimates to the current frame. To remove temporal jitters and outliers from estimates, a fusion network is designed to aggregate depth predictions temporally. Compared with existing methods, \ours{} produces temporally consistent metric depth and is capable of handling dynamic scenes in an online setting. Qualitative results of improved temporal consistency of \ours{} framework are shown in \autoref{fig:frame_frame_jitter}.

For evaluation, we first show empirically that current stereo depth estimation solutions indeed suffer from poor temporal stability. \mli{We quantify such inconsistency with a set of temporal metrics and find that networks with \textit{better} per-frame accuracy may have \textit{worse} temporal consistency}. We then benchmark \ours{} on varied datasets, including synthetic data of rigid \cite{mayer2016large} or deforming \cite{wang2020tartanair} objects, real-world footage of driving \cite{uhrig2017sparsity,menze2015joint}, indoor and medical scenes. \ours{} improves over competing methods in terms of temporal metrics by up to 31\%, and performs on par in terms of per-frame accuracy. The improvement is attributed to the temporal information that our model leverages. We conduct extensive ablation studies of different components of \ours{} and further demonstrate the performance upper bound of our proposed setup empirically, which may motivate future research. \ours{} can run at 25 FPS on modern hardware.

Our contribution can be summarized as follows:
\begin{itemize}
\itemsep 0em
    \item[--] We study an important yet understudied problem for online applications such as augmented reality: \textit{temporal consistency} in-depth estimation from stereo images. We demonstrate that contemporary per-frame solutions suffer from temporal noise.
    \item[--] We present a general framework \ours{} that builds on per-frame stereo depth networks for improved temporal stability. We design a novel \textit{motion} network to accommodate dynamics and a novel \textit{fusion} network to encourage temporal consistency in an online setting. 
    \item[--] We conduct experiments across varied datasets to demonstrate the favorable temporal performance of \ours{} without sacrificing per-frame accuracy. 
\end{itemize}

\begin{figure*}[tb]
    \centering
    \includegraphics[width=\linewidth]{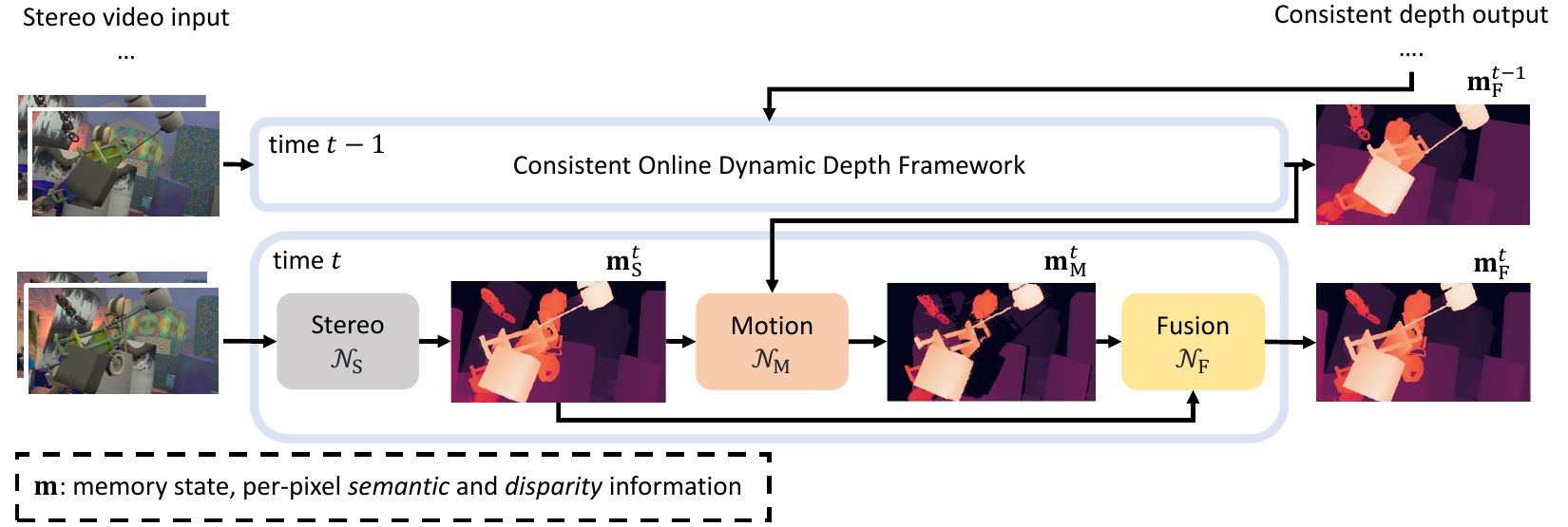}

    \caption{Consistent Online Dynamic Depth (\ours{}) framework has three sub-networks: \textit{stereo} (S), \textit{motion} (M) and \textit{fusion} (F) networks. Each sub-network extracts or updates a memory state $\mathbf{m}$, containing semantic and disparity information. Given an input video sequence, the stereo network extracts an initial estimate $\stereo{m}{t}$ on a per-frame basis. The motion network then aligns the preceding fusion memory state $\fusion{m}{t-1}$ with the current frame, generating the motion memory state $\motion{m}{t}$. A fusion network lastly fuses the two memory states as $\fusion{m}{t}$, containing the temporally consistent disparity $\fusion{d}{t}$. Images from FlyingThings3D dataset \cite{mayer2016large}.}
    \label{fig:overview}
\end{figure*}

\section{Related Work}
\label{sec:related-work}

\textbf{Stereo depth networks} compute disparity between the left and right images to obtain the depth estimate given camera parameters. Classical approaches, such as SGM \cite{hirschmuller2007stereo}, use mutual information to guide disparity estimation. In recent years, different network architectures have been proposed including 3D-convolution-based networks such as PSMNet \cite{chang2018pyramid}, correlation-based networks such as HITNet \cite{tankovich2021hitnet}, hybrid approaches with both 3D convolution and correlation such as GwcNet \cite{guo2019group}, and transformer-based networks such as STTR \cite{li2021revisiting}. \ours{} builds on top of these per-frame methods to improve temporal consistency.

\textbf{Temporally consistent depth networks} aim to produce coherent depth for a video sequence. \textit{Offline} approaches assume no temporal constraints. Some methods \cite{liu2019neural,long2021multi} are only applicable in static scenes while others \cite{luo2020consistent,kopf2021robust} explicitly mask out moving objects and optimize for static background only. Zhang \etal \cite{zhang2021consistent} extend the aforementioned prior methods to dynamic scenes by using additional 3D scene flow estimation. 

\textit{Online} approaches \cite{tananaev2018temporally,zhang2019exploiting} have used recurrent modules to aggregate temporal information. However, these approaches were studied in the context of monocular depth, where errors are dominated by scale inconsistency \cite{kopf2021robust,zhang2021consistent}. Moreover, the recurrent modules do not provide an explicit mechanism of how temporal information is used. \ours{} is instead designed for metric depth from stereo images and provides explicit reasoning between frames in an online setting.

\textbf{Scene flow estimation} seeks to recover inter-frame 3D motion from a set of stereo or RGBD images \cite{ma2019deep,yang2020upgrading,teed2021raft}. While previous algorithms aim at generating accurate scene flow between frames, our work uses the 3D motion as an intermediary such that previous estimates can be aligned with the current frame. Following RAFT3D \cite{teed2021raft}, we predict the inter-frame 3D motion as a per-pixel SE3 transformation.

\textbf{Simultaneous localization and mapping (SLAM)} jointly estimates camera poses and mapping of the scene. Many approaches \cite{newcombe2015dynamicfusion, Innmann2016VolumeDeformRV,Yu2017BodyFusionRC, long2021dssr}, such as DynamicFusion, accumulate information over all past frames. These approaches may include objects that have already exited the scene and are thus not relevant anymore, leading to excess computing. Another common assumption is that only a single or few objects are in the scene, which restricts applicability. \ours{} can be seen as a ``temporally local'' SLAM system where only the immediately preceding frames are considered and no prior information about the scene is assumed. 

\begin{figure*}[tb]
    \centering
    \includegraphics[width=\linewidth]{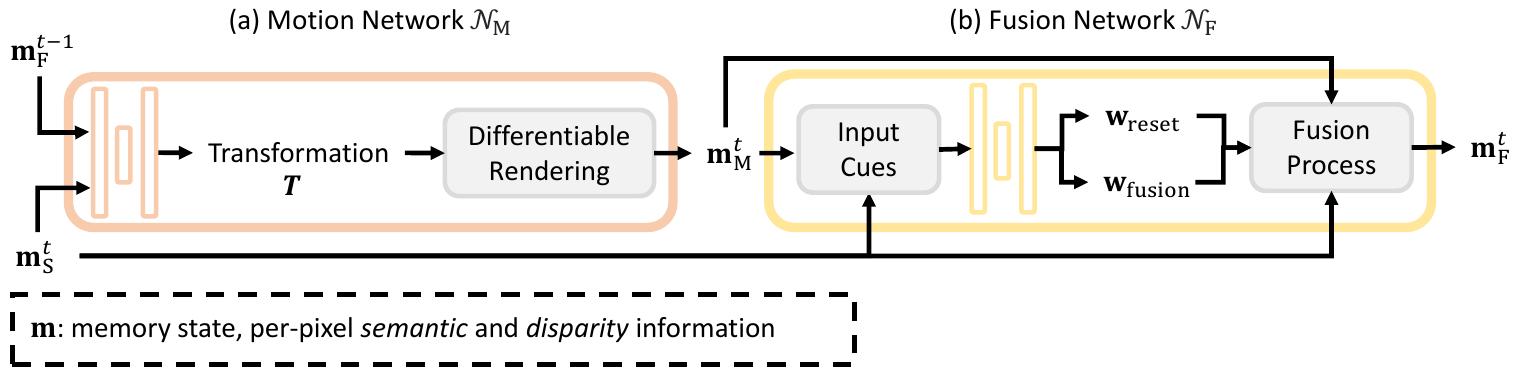}

    \caption{(a) The motion network computes a per-pixel SE3 transformation $\trans$. The transformation $\trans$ is used to align $\fusion{m}{t-1}$ to current stereo state $\stereo{m}{t}$ by differentiable rendering, generating the motion memory state $\motion{m}{t}$. (b) The fusion network first extracts a set of input cues from the memory states $\motion{m}{t}$ and $\stereo{m}{t}$. It then regresses the reset and fusion weights, where $\wreset$ removes outliers and $\wfusion$ aggregates the predictions. The fusion network outputs $\fusion{m}{t}$, containing the temporally consistent disparity estimate $\fusion{d}{t}$.}
    \label{fig:overview_motion_fusion}
\end{figure*}

\section{Consistent Online Dynamic Depth}
The goal of \ours{} is to estimate temporally consistent depth in dynamic scenes in an online setting. A stereo video stream is taken as input, where each image is of dimension $\mathbb{R}^{I_H\times I_W\times 3}$. Let a memory state $\mathbf{m} \in \mathbb{R}^{I_H\times I_W\times (3+C+1)} $ be a combination of per-pixel $3+C$-channel semantic and 1-channel disparity information. \ours{}, as shown in \autoref{fig:overview}, consists of three sub-networks: 

\begin{itemize}
\itemsep 0em
    \item[--] A \textit{stereo} network $\xstereo{\N}$ (\autoref{ssec:stereo_network}) estimates the initial disparity and semantic feature map on a per-frame basis. The semantic information is encoded as feature maps and RGB values.
    \item[--] A \textit{motion} network $\xmotion{\N}$ (\autoref{ssec:motion_network}) accounts for motion across frames by aligning the previous predictions to the current frame. Such motion information is also added to the semantic information for better fusion.
    \item[--] A \textit{fusion} network $\xfusion{\N}$ (\autoref{ssec:fusion_network}) that fuses the disparity estimates across time to promote the temporal consistency in an online setting.
\end{itemize}
We introduce the high-level concepts of each sub-network in the following sections and detail the designs in \aref{app:network_details}.

\subsection{Stereo Network}
\label{ssec:stereo_network}

The objective of the stereo network is to extract an initial estimate of disparity and semantic feature map on a per-frame basis. The stereo network $\N_s$ takes the current stereo images as input and outputs the stereo memory state $\stereo{m}{t}$ at time $t$. Internally, it extracts $\mathbb{R}^{I_H\times I_W\times C}$ semantic feature maps from the stereo images, and computes $\mathbb{R}^{I_H\times I_W\times 1}$ disparity by finding matches between the feature maps. In this paper, we use a recent network, HITNet \cite{tankovich2021hitnet}, as our building block
to estimate per-frame disparity due to its real-time speed and superior performance. In the subsequent sections, we discuss how to extend the stereo network to stabilize the disparity prediction temporally.

\subsection{Motion Network}
\label{ssec:motion_network}
The motion network aligns the previous memory state with that of the current frame. Let $\fusion{m}{t-1}$ denote the preceding memory state from our online consistent depth network at time $t-1$. Our motion network $\xmotion{\N}$, as shown in \autoref{fig:overview_motion_fusion}a, transforms $\fusion{m}{t-1}$ into current frame based on $\stereo{m}{t}$:
\begin{equation*}
    \motion{m}{t} ~\leftarrow~ \xmotion{\mathcal{N}} \big(\fusion{m}{t-1},~ \stereo{m}{t}\big),
\end{equation*}
where $\motion{m}{t}$ is the state from $t-1$ aligned to $t$.

To perform such alignment, the inter-frame motion must be recovered. In a dynamic scene with camera movement and object movement/deformation, the motion prediction needs to be on a per-pixel level. Our motion network builds on top of RAFT3D \cite{teed2021raft} to predict a per-pixel SE3 transformation map $\trans \in \mathbb{SE}(3)^{I_H\times I_W}$. The motion is predicted using a GRU network and a Gauss-Newton optimization mechanism based on matching confidence for $\mathcal{K}$ iterations. Once the motion between frames is recovered, we project the previous memory state $\fusion{m}{t-1}$ to the current frame similar to \cite{li2020robotic} using differentiable rendering \cite{ravi2020pytorch3d}:
\begin{equation}
    \motion{m}{t} ~\leftarrow ~ \pi \Big(\trans\pi^{-1}\big(\fusion{m}{t-1}\big)\Big),
\end{equation}
where $\pi$ and $\pi^{-1}$ are perspective and inverse perspective projection, respectively. Thus, the motion memory state $\motion{m}{t}$ resides in the current camera coordinate frame and has pixel-wise correspondence with the current prediction, which enables temporal aggregation of disparity. Additionally, we estimate a binary visibility mask by identifying the regions in the current frame that is not visible in the previous one. We also compute the confidence of motion via Sigmoid and compute the motion magnitude as the L2 norm of the scene flow. These information are added to the memory state $\motion{m}{t}$ for the fusion network to adaptively fuse the predictions. We detail differences between our motion network and RAFT3D and provide a quantitative comparison in \aref{app:motion_network}.

\subsection{Fusion Network}
\label{ssec:fusion_network}
The objective of the fusion network (\autoref{fig:overview_motion_fusion}b) is to promote temporal consistency by aggregating the disparities of the motion and stereo memory states. The output of the fusion network is the fusion memory state $\fusion{m}{t}$:
\begin{equation}
    \fusion{m}{t} ~ \leftarrow ~ \xfusion{\mathcal{N}} \big(\motion{m}{t},~ \stereo{m}{t} \big),
\end{equation}
where $\xfusion{\N}$ is the fusion network and $\fusion{m}{t}$ contains the temporally consistent disparity estimate $\fusion{d}{t}$. We first discuss the fusion process (\autoref{sssec:fusion}) and then cover the set of cues extracted from the memory states (\autoref{sssec:input_cues}) to guide such fusion process.

\subsubsection{Fusion Process} 
\label{sssec:fusion}
The temporally consistent disparity $\fusion{d}{t}$ is obtained by fusing the aligned and current disparity estimates. Let $\motion{d}{t}$ be the disparity from the motion network and $\stereo{d}{t}$ be the disparity from the stereo network. The fusion network computes a reset weight $\wreset$ and a fusion weight $\wfusion$ both of dimension $\mathbb{R}^{I_H\times I_W}$. The fusion process of disparity estimates is formulated as:
\begin{equation}
    \fusion{d}{t} = \big(1-\mathbf{w}_{\text{reset}} \, \mathbf{w}_{\text{fusion}}\big) \, \stereo{d}{t} +  \mathbf{w}_{\text{reset}} \, \mathbf{w}_{\text{fusion}} \, \motion{d}{t}\,.
    \label{eqn:fusion}
\end{equation}
The fusion memory state $\fusion{m}{t}$ is thus formed by the fused disparity $\fusion{d}{t}$ and semantic features extracted by the stereo network.

The intuition behind the fusion process is two-fold. First, it filters out outliers using $\wreset$. The outliers can be either induced by inaccurate disparity or motion predictions. In our work, $\wreset$ is supervised to identify outliers whose errors are larger than the other disparity estimate by a threshold of $\tau_\text{reset}$. Second, the fusion process encourages temporal consistency by fusing current disparity prediction with reliable predictions propagated from the previous frame using $\wfusion$. When disparity estimates are considered to be equally reliable within a threshold $\tau_\text{fusion}$, the fusion network aggregates them with a regressed value between 0 and 1. As reset weights should already reject the most significant outliers, we set $\tau_\text{fusion} < \tau_\text{reset}$.

\subsubsection{Input Cues}
\label{sssec:input_cues}
To determine the reset and fusion weights, we collect a set of input cues from $\motion{m}{t}$ and $\stereo{m}{t}$. We find explicit input cues are advantageous over the channel-wise concatenation of the two memory states.

First, the \textit{\textbf{disparity confidence}} of $\stereo{d}{t}$ and $\motion{d}{t}$ are computed. As disparities are computed mainly based on appearance similarity between the left and right images, we approximate the confidence of disparity prediction by computing the $\ell_1$ distance between the left and right features extracted from the stereo network. For robustness against matching ambiguity, we additionally offset each disparity estimate by $-1$ and $1$ to collect local confidence information, forming a 3-channel confidence feature.

\begin{figure}[bt]
    \centering
    \includegraphics[width=0.8\linewidth]{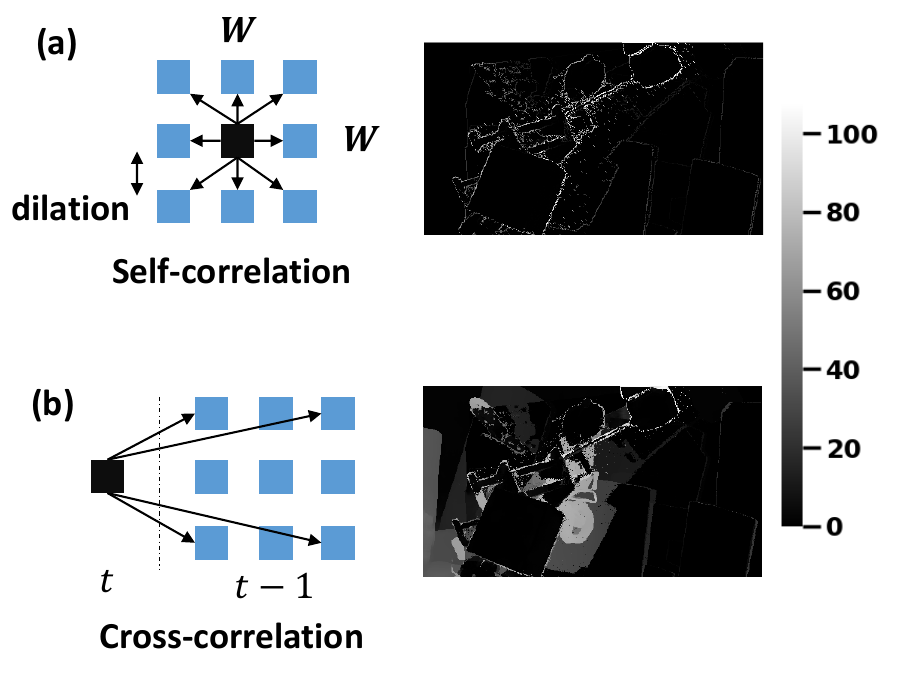}
    \caption{The pixel-to-patch correlation mechanisms and example visualizations of a single channel of disparity correlations. Black pixels: source pixels. Blue pixels: correlated pixels. We note that, qualitatively, self-correlation identifies local discontinuities and cross-correlation identifies inter-frame disagreements. Images from FlyingThings3D dataset \cite{mayer2016large}.}
    \label{fig:attn}
\end{figure}

However, in the case of stereo occlusion, \ie regions that are only visible in the left but not the right image, the disparity confidence based on appearance similarity becomes ill-posed. Thus, we additionally use the local smoothness information as a cue. We implement a pixel-to-patch \textit{\textbf{self-correlation}} (\autoref{fig:attn}a) to approximate the local smoothness information, which computes the correlation between a pixel and its neighboring pixels in a local patch of size $W\times W$, forming a $\mathbb{R}^{I_H \times I_W \times (W^2-1)}$ correlation feature. Dilation may be used to increase the receptive field. We apply the pixel-to-patch self-correlation for both disparity and semantic features to acquire local disparity and appearance smoothness. 

In the case of inaccurate motion predictions, the aligned memory state may contain wrong cross-frame correspondences. Therefore, the predictions in these regions need to be discarded as outliers. To promote such outlier rejection, the fusion network applies a pixel-to-patch \textit{\textbf{cross-correlation}} (\autoref{fig:attn}b) to evaluate the cross-frame disparity and appearance similarities. In cross-correlation, each pixel attends to a local $W\times W$ patch of the previous image centered at the same pixel location after motion correction, forming a $\mathbb{R}^{I_H \times I_W \times W^2}$ correlation feature. In our implementation, we use the $\ell_1$ distance for disparity and dot-products for appearance correlation.

Lastly, we observe that when the inter-frame motion is large, the motion estimate is less reliable and may thus result in wrong cross-frame correspondences. Therefore, we provide the flow magnitude and confidence as a motion cue. We additionally use the visibility mask from the projection process to identify the invalid regions and provide the semantic features for context information. 

\subsection{Supervision}
\label{ssec:losses}
\textbf{Stereo and Motion Network} We supervise the stereo network on the per-frame disparity estimate $\stereo{d}{t}$ against the ground truth following \cite{tankovich2021hitnet}. We supervise the motion network on the transformation prediction $\trans$, with a loss imposed on its projected scene flow against the ground truth following \cite{teed2021raft}.

\textbf{Fusion Network} We supervise the fusion network to promote temporal consistency of disparity estimates. We note that optimizing for disparity changes only may not be ideal, as errors in the previous frame will propagate to the current frame even if the predicted disparity change is correct. Therefore, we impose losses on the disparity prediction $\fusion{d}{t}$, and predicted weights $\wreset, \wfusion$. 

We supervise the predicted disparity $\fusion{d}{t}$ against the ground truth using Huber loss \cite{huber1992robust}, denoted as the disparity loss $\ell_\text{disp}$. 

Further, we supervise the reset weights $\wreset$ such that they reject the worse prediction between the stereo and motion network disparity estimates. Let $e_\text{M}=|\motion{d}{t} - \mathbf{d}_\text{gt}^t|$ be the error of the motion disparity and $e_\text{S}=|\stereo{d}{t} - \mathbf{d}_\text{gt}^t|$ be the error of the stereo disparity against the ground truth $\mathbf{d}_\text{gt}^t$. Because $\wreset$ rejects the motion disparity estimate $\motion{d}{t}$ when its value is zero (\autoref{eqn:fusion}), we impose a loss such that it favors zero when $e_\text{M}$ is worse and favors one when $e_\text{M}$ is better. Thus, we have
\begin{equation*}
    \ell_{\text{reset}} = \begin{cases}
        \mathbf{w}_{\text{reset}}, &  1)~\text{if} \ e_\text{M} > e_\text{S} + \tau_\text{reset}, \\
        1 - \mathbf{w}_{\text{reset}}, & 2)~\text{if} \ e_\text{M} < e_\text{S} - \tau_\text{reset}, \\
        0, & 3)~\text{otherwise},
    \end{cases}
\end{equation*}
where $e_\text{M}$ is worse in condition 1) while $e_\text{M}$ is better in condition 2). Otherwise, the loss is zero as shown in condition 3).

The fusion weights $\wfusion$ are supervised such that they aggregate the past two disparity estimates correctly: 
\begin{equation*}
    \ell_{\text{fusion}} = \begin{cases}
        \mathbf{w}_{\text{fusion}}, & 1)~\text{if} \ e_\text{M} > e_\text{S} + \tau_\text{fusion}, \\
        1 - \mathbf{w}_{\text{fusion}}, & 2)~\text{if} \ e_\text{M} < e_\text{S} - \tau_\text{fusion},  \\
        \alpha_\text{reg} \cdot |\mathbf{w}_{\text{fusion}} - 0.5|, & 3)~\text{otherwise}.
    \end{cases} \label{eqn:fusion_loss}
\end{equation*}
Different from the reset weights, the fusion weights are not only trained to identify the better estimate as shown in conditions 1) and 2), but also trained with an additional regularization term such that the fusion weights are around 0.5 when both estimates are considered equally good as shown in condition 3).

The final loss for fusion network training is computed as:
\begin{equation}
    \xfusion{\ell} = \alpha_{\text{disp}}\ell_{\text{disp}} + \alpha_{\text{fusion}}\ell_{\text{fusion}} + \alpha_{\text{reset}}\ell_{\text{reset}}\,, \\
\end{equation} 
which is the weighted sum of $\ell_\text{disp},\ell_\text{fusion}$ and $\ell_\text{reset}$.

\section{Experimental Setup}
\label{sec:experimental_setup}

\subsection{Implementation Details}
We use a batch of 8 and Adam \cite{kingma2014adam} as the optimizer on Nvidia V100 GPUs. Following \cite{teed2021raft}, we use a linearly decayed learning rate of $2\mathrm{e}{-4}$ for pre-training and $2\mathrm{e}{-5}$ for fine-tuning. We pre-train motion and fusion networks for 25000 and 12500 steps, and halve the steps during fine-tuning. We perform $\mathcal{K}=1$ steps of incremental updates in the motion network when datasets contain small motion (\eg TartanAir \cite{wang2020tartanair}) and $\mathcal{K}=16$ otherwise (\eg FlyingThings3D \cite{mayer2016large}, KITTI Depth \cite{uhrig2017sparsity} and KITTI 2015 \cite{menze2015joint}). In the fusion network, we use a patch size $W=3$ with dilation of 2 for pixel-to-patch correlation to increase the receptive field. By default, we set $\tau_\text{reset}=5$, $\tau_\text{fusion}=1$, $\alpha_\text{reg}=0.2$, and $\alpha_{\text{disp}}=\alpha_{\text{fusion}}=\alpha_{\text{reset}}=1$. Due to the sparsity of ground truth in KITTI datasets, supervising the fusion weights can be ill-posed as only a few pixels are aligned across time. We set $\alpha_{\text{fusion}}=\alpha_{\text{reset}}=0$.

\subsection{Metrics}
\label{ssec:metric_and_eval}
We propose to quantify the temporal inconsistency by a temporal end-point-error ($\tepe$) metric and the relative error ($\teper$) given cross-frame correspondences:
\begin{equation}
    \tepe = |\Delta d - \Delta d_\text{gt}|\,,\,\, \teper = \frac{\tepe}{|\Delta d_\text{gt}| + \epsilon}\,,
\end{equation}
where $\Delta d, \Delta d_\text{gt}$ are \textit{signed} disparity change and $\epsilon = 1\mathrm{e}{-3}$ avoids division by zero. Intuitively, $\tepe$ and $\teper$ reflects the absolute and relative error between predicted and ground truth depth motion between two time points.  $\tepe$ is generally proportional to the ground truth magnitude and thus better reflects consistencies in pixels with large motion. $\teper$ better captures the consistencies of static pixels due to the $1/\epsilon$ weight. We also report threshold metrics of 3px for $\tepe$ and 100\% for $\teper$ ($\dmetricsup{3px}{t}$, $\dmetricsup{100\%}{t}$). Temporal metrics themselves can be limited as a network can be temporally consistent but wrong. Therefore, we also report the per-pixel disparity error using EPE and threshold metric of 3px ($\dmetric{3px}$). We exclude pixels with extreme scene flow ($>$ 210px) or disparity ($<$1px or $>$210px) following \cite{teed2021raft} as they are generally outliers in our intended application. For all metrics, lower is better.

\section{Results and Discussion}
\label{sec:experimental_results}
We first show quantitatively that current stereo depth estimation solutions suffer from poor temporal consistency  (\autoref{ssec:temporal_consistency_eval}). We then show that \ours{} improves upon per-frame stereo networks and outperforms competing approaches across varied datasets, \mli{\textit{sharing the same stereo network without the need of re-training or fine-tuning}} (\autoref{ssec:flyingthings3d_result}--\autoref{ssec:benchmark}). We lastly present ablation studies (\autoref{ssec:ablation}) and inference time (\autoref{ssec:inference}) to characterize \ours{}.

\setlength\tabcolsep{0.7em}
\begin{table}[b]
\centering
\caption{Temporal and per-pixel metrics of contemporary approaches evaluated on the FlyingThings3D dataset \cite{mayer2016large} with the official checkpoints when provided. $\dagger$: Classical approach. $\ddagger$: Non-occluded regions only. We use HITNet \cite{tankovich2021hitnet} as the stereo network due to its real-time speed and superior performance.}
\label{tab:flyingthings3d}
\resizebox{\linewidth}{!}{%
\begin{tabular}{c|cc|cc|cc}
\toprule
 & $\tepe$ & $\dmetricsup{3px}{t}$ & $\teper$ & $\dmetricsup{100\%}{t}$ & EPE& $\dmetric{3px}$  \\ \midrule
SGM \cite{hirschmuller2007stereo} $\dagger$ & 2.355 &  0.065 & 78.482 & 0.591 & 2.965 & 0.090 \\
STTR \cite{li2021revisiting} $\ddagger$ & 0.482 & 0.014 & 11.434 & 0.374 & 0.449 & 0.014 \\ \midrule
PSMNet \cite{chang2018pyramid} & 1.371 & 0.056 & 35.136 & 0.466 & 1.079 & 0.045 \\ 
GwcNet & 0.959 & 0.041 & 22.598 & 0.409 & 0.752 & 0.032 \\ 
HITNet \cite{tankovich2021hitnet} & 0.812 & 0.040 & 16.840 & 0.291 & 0.607 & 0.030 
\end{tabular}%
}
\end{table}

\subsection{Temporal Consistency Evaluation}
\label{ssec:temporal_consistency_eval}
We examine the temporal consistency of stereo depth techniques of different designs that operate on a per-frame basis. We use FlyingThings3D finalpass \cite{mayer2016large} dataset as it is commonly used for training stereo networks. 

\textbf{Results} As shown in \autoref{tab:flyingthings3d}, all considered approaches have large $\tepe$ and $\teper$, with $\tepe$ often larger than EPE and more than 14 times the ground truth disparity change as implied by $\teper$. This suggests that the considered approaches suffer from poor temporal stability despite good per-pixel accuracy. A qualitative visualization is shown in \autoref{fig:frame_frame_jitter}. We show the resulting artifacts in the video supplement to further illustrate the case.

\subsection{Pre-training on FlyingThings3D}
\label{ssec:flyingthings3d_result}
We first demonstrate that \ours{} improves the temporal stability of per-frame networks by using the pre-trained HITNet \cite{tankovich2021hitnet} as our stereo network and freezing its parameters during training for a fair comparison. We follow the official split of FlyingThings3D. \autoref{tab:flyingthings3d_ours} summarizes the result.

\setlength\tabcolsep{0.5em}
\begin{table}[t]
\centering
\caption{Results on the FlyingThings3D dataset \cite{mayer2016large}.}
\label{tab:flyingthings3d_ours}
\resizebox{\linewidth}{!}{%
\begin{tabular}{c|cc|cc|cc}
\toprule
 & $\tepe$ & $\dmetricsup{3px}{t}$ & $\teper$ & $\dmetricsup{100\%}{t}$ & EPE & $\dmetric{3px}$  \\ \midrule
Stereo \cite{tankovich2021hitnet} & 0.812 & 0.040 & 16.840 & 0.291 & 0.607 & 0.030 \\
Motion & 0.875 & 0.036 & 24.533 & 0.390  & 0.777 & 0.030 \\ \hline
Kalman filter \cite{kalman1960new} & 0.793 & 0.040 & 15.843 & 0.230 & 0.610 & 0.030 \\
\textbf{\ours{} (Ours)} & \bf 0.741 & \bf 0.034 & \bf 15.205 & \bf 0.214 & \bf 0.595 & \bf 0.028
\end{tabular}%
}
\end{table}

\textbf{Results} \mli{To ensure temporal consistency, one naive way is to only forward the past information to the current frame. Thus, we compare the results of the per-frame prediction $\stereo{d}{t}$ of the stereo network with the aligned preceding prediction $\motion{d}{t}$ of the motion network. To evaluate fairly, we fill occluded regions that are not observable in the past with $\stereo{d}{t}$. As shown in \autoref{tab:flyingthings3d_ours}, the motion estimate $\motion{d}{t}$ is worse than stereo estimate $\stereo{d}{t}$ due to outliers caused by wrong motion predictions as shown blue in \cref{fig:qual_vis_flythings3d}, suggesting that only forwarding past information is not feasible. Nonetheless, $\motion{d}{t}$ performs on par with $\stereo{d}{t}$ in most of the cases (white) and can even mitigate errors of $\stereo{d}{t}$ that is hard to predict in the current frame (red). Thus, techniques to adaptively handle these cases can greatly reduce temporal noise.}

\mli{While Kalman filter \cite{kalman1960new} successfully improves the temporal consistency by combining the two outputs, it leads to worse EPE. The result indicates that temporal consistency is indeed a \textit{different} problem from per-frame accuracy. In contrast, \ours{} performs better in all metrics, which suggests that \ours{} achieves better stability across frames by pushing predictions toward the ground truth instead of propagating errors in time. We further show that our pipeline applies to other stereo networks in \aref{app:other-stereo}.}

\setlength\tabcolsep{0.6em}
\begin{table*}[bt]
\begin{minipage}{.4\linewidth}
    \centering
    \captionof{figure}{Temporal consistency comparison between stereo $\stereo{d}{t}$ and motion $\motion{d}{t}$ predictions. While $\motion{d}{t}$ contains outliers (blue), it can also mitigate errors of $\stereo{d}{t}$ (red). Our fusion network learns to adaptively fuse the estimates. Images from FlyingThings3D dataset \cite{mayer2016large}.} 
    \includegraphics[width=0.9\linewidth]{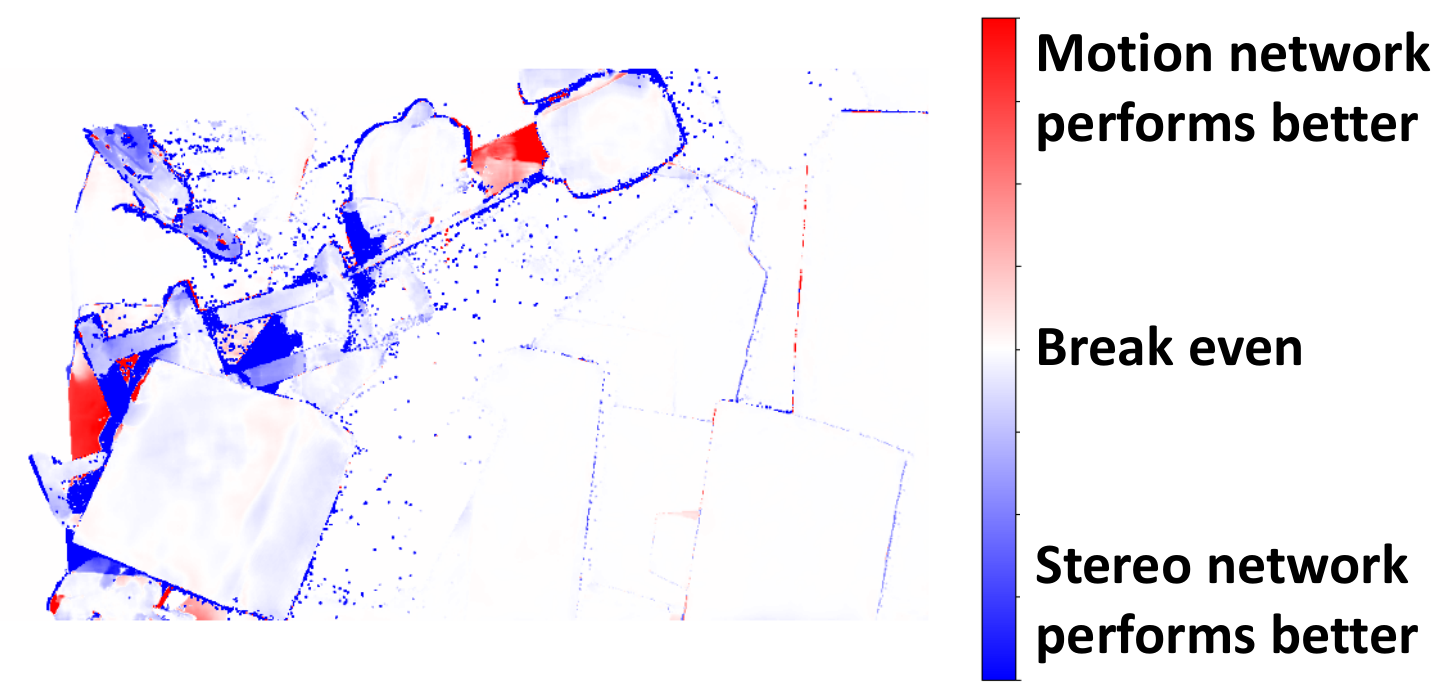}
    \label{fig:qual_vis_flythings3d}
\end{minipage}
\hspace{1em}
\begin{minipage}{0.57\linewidth}
    \centering

    \caption{Results on TartanAir \cite{wang2020tartanair}, KITTI Depth \cite{uhrig2017sparsity}, KITTI 2015 \cite{menze2015joint}.}
    \label{tab:fine-tuning}

    \medskip
    
    \resizebox{\linewidth}{!}{%
    \begin{tabular}{c|c|cc|cc|cc}
    \toprule
         \multicolumn{2}{c|}{} & $\tepe $ & $\dmetricsup{3px}{t}$ & $\teper $ & $\dmetricsup{100\%}{t} $ & EPE & $\dmetric{3px} $ \\ \midrule
    \multirow{3}{*}{\shortstack{TartanAir \\ dataset \cite{wang2020tartanair}}} & Stereo \cite{tankovich2021hitnet} & 0.876 & 0.053 & 9.039 & 0.339 & 0.934 & 0.055  \\ 
    & Kalman Filter \cite{kalman1960new} & 0.829 & 0.053 & 7.501 & 0.252 & 0.935 & 0.055 \\ 
    & \textbf{\ours{} (Ours)} & \bf 0.751 & \bf 0.045 & \bf 6.206 & \bf 0.240 & \bf 0.904 & \bf 0.053
    \\ \midrule
    \multirow{3}{*}{\shortstack{KITTI Depth \\ dataset \cite{uhrig2017sparsity}}} & Stereo \cite{tankovich2021hitnet} & 0.289 & \bf 0.001 & 3.630 & 0.156 & 0.423 & 0.004  \\ 
    & Kalman Filter \cite{kalman1960new} & 0.278 &  \bf 0.001 & \bf 2.615 & \bf 0.125 & 0.431 & 0.005  \\ 
    & \textbf{\ours{} (Ours)} & \bf 0.258 & \bf 0.001 & 2.764 & 0.132  & \bf 0.418 & \bf 0.003
    \\ \midrule
    \multirow{3}{*}{\shortstack{KITTI 2015 \\ dataset \cite{menze2015joint}}} & Stereo \cite{tankovich2021hitnet} & 0.570 & 0.026 & 10.672 & 0.126 & \multirow{3}{*}{0.811} & \multirow{3}{*}{0.033} \\ 
    & Kalman Filter \cite{kalman1960new} & 0.533 & 0.026 & 9.668 & 0.117 &  & 
     \\ 
    & \textbf{\ours{} (Ours)} & \bf 0.507 & \bf 0.022 & \bf 8.740 & \bf 0.112 & & 
    \end{tabular}
    }
\end{minipage}
\end{table*}

\subsection{Benchmark Results}
\label{ssec:benchmark}
We benchmark \ours{} across varied datasets. We first fine-tune the stereo network on each dataset to ensure good per-frame accuracy and keep it frozen for a fair comparison. We then fine-tune the motion and fusion networks, using the output of the stereo network as input. We summarize results in \autoref{tab:fine-tuning}, provide end-to-end results in \aref{app:e2e_training} and zero-shot results in \aref{app:zero-shot}.

\textbf{Dataset} TartanAir \cite{wang2020tartanair} is a synthetic dataset with simulated drone motions in different scenes. We use 15 scenes (219434 images) for training, 1 scene for validation (6607 images), and 1 scene (5915 images) for testing. 

KITTI Depth \cite{uhrig2017sparsity} has footage of real-world driving scenes, where ground truth depth is acquired from LiDAR. We follow the official split and train \ours{} on 57 scenes (38331 images), validate on 1 scene (1100 images), and test on 13 scenes (3426 images). We use pseudo ground truth information inferred from an off-the-shelf optical flow network \cite{teed2020raft} trained on KITTI 2015.

KITTI 2015 \cite{menze2015joint} is a subset of KITTI Depth with 200 temporal image pairs. The data is complementary to KITTI Depth as ground truth optical flow information is provided, however, the long-term temporal consistency is not captured as there are only two frames for each scene. We train on 160 image pairs, validate on 20 image pairs, and test on 20 image pairs. Given the small dataset size, we perform five-fold cross-validation experiments and report the average results.

\textbf{Results} We find that \ours{} consistently outperforms the per-frame stereo network \cite{tankovich2021hitnet} across all metrics similar to findings in \autoref{ssec:flyingthings3d_result}. The $\teper$ is improved by up to 31\%, from 9.039 to 6.206 in the TartanAir dataset. \mli{Compared to the Kalman filter, \ours{} leads to smaller EPE in contrast to the Kalman filter, indicating both improved temporal and per-pixel performance. This again demonstrates the improved temporal performance does \textit{NOT} guarantee improved per-pixel accuracy.} We note that all settings in KITTI 2015 have the same EPE and $\dmetric{3px}$, because we do not have ground truth information to evaluate the per-pixel accuracy of the fusion result of the second frame. More qualitative visualizations can be found in video supplement and \aref{app:qualitative_vis}.

\subsection{Ablation Experiments}
\label{ssec:ablation}
We conduct experiments on the FlyingThings3D dataset \cite{mayer2016large} to ablate the effectiveness of different sub-components.

\subsubsection{Fusion Network}
\label{sssec:ablation_fusion}
We summarize the key ablation experiments of the fusion network in \autoref{tab:ablation_fusion} and provide additional results in \aref{app:additional_fusion_ablation}.

\setlength\tabcolsep{.5em}
\begin{table}[bt]
\centering
\caption{Ablation studies of fusion network. \underline{Underline}: base setting.}
\label{tab:ablation_fusion}
\resizebox{\linewidth}{!}{%
\begin{tabular}{cc|cc|cc|cc}
\toprule
\multicolumn{2}{c|}{} & $\tepe $ & $\dmetricsup{3px}{t} $ & $\teper $ & $\dmetricsup{100\%}{t} $ & EPE & $\dmetric{3px} $  \\ \midrule
a) Reset & $\times$ & 0.783 & 0.036 & 15.953 & 0.217 & 0.618 & 0.030 \\ 
weight & \underline{\checkmark} & \bf 0.756 & \bf 0.035 & \bf 15.013 & \bf 0.211 & \bf 0.604 & \bf 0.029 \\ \midrule
\multirow{3}{*}{\shortstack{b) Fusion \\ input \\ cues}} & +FL & 0.763 & 0.035 & 15.103 & 0.211 & \bf 0.604 & 0.029 \\ 
& +V & 0.758 & 0.035 & 15.082 & \bf 0.210 & 0.605 & 0.029 \\ 
& \underline{+SM} & \bf 0.756 & 0.035 & \bf 15.013 & 0.211 & \bf 0.604 & 0.029 \\ \midrule
c) Training & \underline{2} & 0.756 & 0.035 & 15.013 & \bf 0.211 & 0.604 & 0.029 \\ 
sequence & 3 & 0.753 & 0.035 & \bf 14.942 & \bf 0.211 & 0.600 & 0.029 \\
length & 4 & \bf 0.741 & \bf 0.034 & 15.205 & 0.214 & \bf 0.595 & \bf 0.028
\end{tabular}%
}
\end{table}

\textit{a) Reset weights:} In theory, only the fusion weight $\wfusion$ is needed between two estimates as it can reject outliers by estimating extreme values (\eg, 0 or 1). However, we find that predicting additional reset weights $\wreset$ improves performance across all metrics. This may be attributed to the difference in supervision, where reset weights $\wreset$ are trained for outlier detection, while fusion weights $\wfusion$ are trained for aggregation. 

\textit{b) Fusion input cues:} Other than the disparity confidence, self- and cross-correlation, we incrementally add the flow confidence/magnitude (+FL), visibility mask (+V), and semantic feature map (+SM) to the fusion networks. We find that the metrics improve marginally with additional inputs, especially for $\tepe$.

\textit{c) Training sequence length:} During training, by default we train with sequences with a length of two frames, where the second frame takes the stereo outputs of the first frame as input. However, during the inference process, the preceding outputs are from the fusion network. Thus, to better approximate the inference process, we further extend the sequence length to three and four. We find that the fusion network consistently benefits from the increasing training sequence length. However, this elongates training time proportionally.

\subsubsection{Empirical Best Case}
\label{sssec:empirical_best}
We provide an empirical study of the ``best case'' of motion or fusion network in \autoref{tab:empirical_best}. For empirical best motion network $\xmotion{\N}$, we use ground truth scene flow and set flow confidence to one. For empirical best fusion network $\xfusion{\N}$, we pick the better disparity estimates pixel-wise between the stereo $\stereo{d}{t}$ and motion $\motion{d}{t}$ estimates given ground truth disparity. We find that perfect motion leads to a substantial reduction in $\teper$ while perfect fusion leads to a substantial reduction in $\tepe$. In both cases, $\epe$ also improves. While \ours{} performs well against competing approaches, advancing the motion or fusion networks has the potential to substantially improve temporal consistency.

\setlength\tabcolsep{.5em}
\begin{table}[tb]
\centering
\caption{Ablation study of the empirical best case using ground truth information.}
\label{tab:empirical_best}
\resizebox{\linewidth}{!}{%
\begin{tabular}{cc|cc|cc|cc}
\toprule
\multicolumn{2}{c|}{} & $\tepe $ & $\dmetricsup{3px}{t} $ & $\teper $ & $\dmetricsup{100\%}{t} $ & EPE & $\dmetric{3px} $  \\ \midrule
\multicolumn{2}{c|}{\bf \ours{} (Ours)} & 0.741 & 0.034 & 15.205 & 0.214 & 0.595 & 0.028 \\
\multicolumn{2}{c|}{Empirical best $\xmotion{\N}$} & 0.879 & 0.043 & \bf 5.401 & \bf 0.125 & 0.571 & 0.027 \\
\multicolumn{2}{c|}{Empirical best $\xfusion{\N}$} & \bf 0.529 & \bf 0.025 & 9.936 & 0.214 & \bf 0.455 & \bf 0.021
\end{tabular}%
}
\end{table}

\subsection{Inference Speed and Number of Parameters}
\label{ssec:inference}
The inference speed of \ours{} on images of resolution 640$\times$480 with $\mathcal{K}=1$ is 25 FPS (stereo 26ms, motion 13ms, fusion 0.3ms) on an Nvidia Titan RTX GPU. The total number of parameters is 9.3M (stereo 0.6M, motion 8.5M, and fusion 0.2M). Compared to the stereo network, the overhead is mainly introduced by the motion network. 

\subsection{Limitations}
\label{ssec:limitation}
While \ours{} outperforms competing methods across datasets, we recognize that there is still a gap between \ours{} and the empirical best cases in \autoref{sssec:empirical_best}. Furthermore, \ours{} cannot correct errors when both current and previous frame estimates are wrong, as the weights between estimates in the fusion process are bounded to $(0,1)$. Lastly, we design \ours{} to look at the immediately preceding frame only. Exploiting more past information may potentially lead to performance improvement.

\section{Conclusions}
We present a general framework to produce temporally consistent depth estimation for dynamic scenes in an online setting. \ours{} builds on contemporary per-frame stereo depth approaches and shows superior performance across different benchmarks, both in terms of temporal metrics and per-pixel accuracy. Future work may extend upon our motion or fusion network for better performance and extend to multiple past frames. 

{\small
\bibliographystyle{ieee_fullname}
\bibliography{egbib}
}

\newpage

\textbf{Acknowledgements} We thank Johannes Kopf and Michael Zollhoefer for helpful discussions, and Vladimir Tankovich for HITNet implementation. We thank anonymous reviewers and meta-reviewers for their feedback on the paper. This work was done during an internship at Reality Labs, Meta Inc and funded in part by NIDCD K08 Grant DC019708.

\appendix
\section{Network Details}
\label{app:network_details}
Our code base is built upon MMSegmentation \cite{mmseg2020}.

\subsection{Stereo Network}
We use HITNet \cite{tankovich2021hitnet} as our per-frame stereo network to extract disparity and features on a per-frame basis. HITNet builds an initial cost volume with a pre-specified disparity range. In our implementation, we set the maximum disparity to be 320 following the original implementation. The channel size of extracted features is 24, which is used in the fusion network to evaluate disparity confidence. The memory state generated by the stereo network includes the frame-wise estimated disparity, the feature of left and right images at the 1/4 resolution for disparity confidence computation, and the left RGB image. 

Our HITNet code is adapted from open-sourced implementation\footnote{https://github.com/MJITG/PyTorch-HITNet-Hierarchical-Iterative-Tile-Refinement-Network-for-Real-time-Stereo-Matching}\footnote{https://github.com/meteorshowers/X-StereoLab} in PyTorch \cite{paszke2019pytorch} and confirmed by the authors of HITNet. We note that the HITNet adapted is based on Table 6 of \cite{tankovich2021hitnet} instead of the HITNet XL or HITNet L in Table 1 of \cite{tankovich2021hitnet}. As HITNet works across different datasets (Table 2-4 in \cite{tankovich2021hitnet}), we have chosen HITNet as our stereo network. As described in Sect.4.2, we remove pixels with extreme scene flow ($>$210px) and disparity ($<$1px or $>$210px). These removed pixels correspond to unrealistically fast, far, or close objects during simulation, which are not considered in our intended applications such as augmented reality. In comparison, \cite{tankovich2021hitnet} removes pixels with disparity $>$192px. Thus, the reported per-pixel accuracy performance differs slightly on FlyingThings3D.

\subsection{Motion Network}
\label{app:motion_network}
Our motion network builds on top of RAFT3D \cite{teed2021raft}. Specifically, we adapt the inter-frame transformation $\trans$ estimation process. We first provide the necessary background of the estimation process and detail the difference between our motion network and RAFT3D next paragraph. Let $\mathbf{p}$ be a location in pixel coordinates and $\mathbf{P}$ be a location in Cartesian coordinates. RAFT3D uses a context extractor to provide semantic information for object grouping, and a feature extractor for cross-frame correspondence matching. A piece-wise rigid constraint is realized by grouping pixels based on their motion and appearance similarities and enforcing the motion within a group to be the same. The GRU module takes the image context feature, correlation information, estimated transformation, and corresponding scene flow as input. The GRU then makes corrections $\Delta \mathbf{s}^k$ to the scene flow, extracts a set of rigid-embeddings $\mathcal{V}$, and associated confidence of estimates $\mathbf{w}$. A differentiable Gauss-Newton (GN) optimization step is performed to update the transformation based on the residual error (\cref{eqn:gauss_netwon_step}). The update scheme is iterative for $\mathcal{K}$ steps, with the transformation $\trans^0$ and flow $\mathbf{s}^0$ initialized to identity and zero respectively. Given the \textit{i}-th pixel at $t-1$ and its local $m$ neighborhood pixels $\mathcal{N}_{m}(i)$, the \textit{k}-th iteration residual error to be minimized is: 
\begin{align}
E_i^k(\zeta_i) &= \sum_{j \in \mathcal{N}_{m}(i)} \mathcal{V}_{ij} \| \mathbf{p}_j +\mathbf{s}_j^{k} - \pi(e^{\zeta^k_{i}}\trans_i^{k-1}\mathbf{P}_j) \|_{2,\mathbf{w}_j} 
\label{eqn:gauss_netwon_step} \\
\mathbf{s}^{k}_j &= \pi(\trans_i^{k-1}\mathbf{P}_j) - \mathbf{p}_j+ \Delta \mathbf{s}^k_j \\
\mathcal{V}_{ij} &= \|\mathcal{V}_i - \mathcal{V}_j\|_2
\end{align}
\noindent where $e^{\zeta^k_i}$ is the incremental motion on the SE3 manifold to be made to the previous transformation $\trans^{k-1}_i$, $\mathbf{s}^{k}_j $ is the raw predicted scene flow, $\|\cdot\|_{2,\mathbf{w}}$ is weighted $\ell_2$ norm, $\mathbf{w}_j$ is the flow confidence, and $\mathcal{V}_{ij}$ is the $\ell_2$ distance of rigid-embeddings. The updated transformation is $\trans^k = e^{\zeta^k}\trans^{k-1}$. The final scene flow-induced transformation from transformation prediction $\trans$ in image coordinate can be computed as:
\begin{equation}
    \mathbf{s}_{\trans} = \pi(\trans \pi^{-1}(\mathbf{p})) - \mathbf{p}
\end{equation}
The magnitude $\mathbf{s}_{\trans}$ and confidence $\mathbf{w}_j$ of the flow is used later in fusion network.

\begin{figure}[tb]
    \centering
    \includegraphics[width=\linewidth]{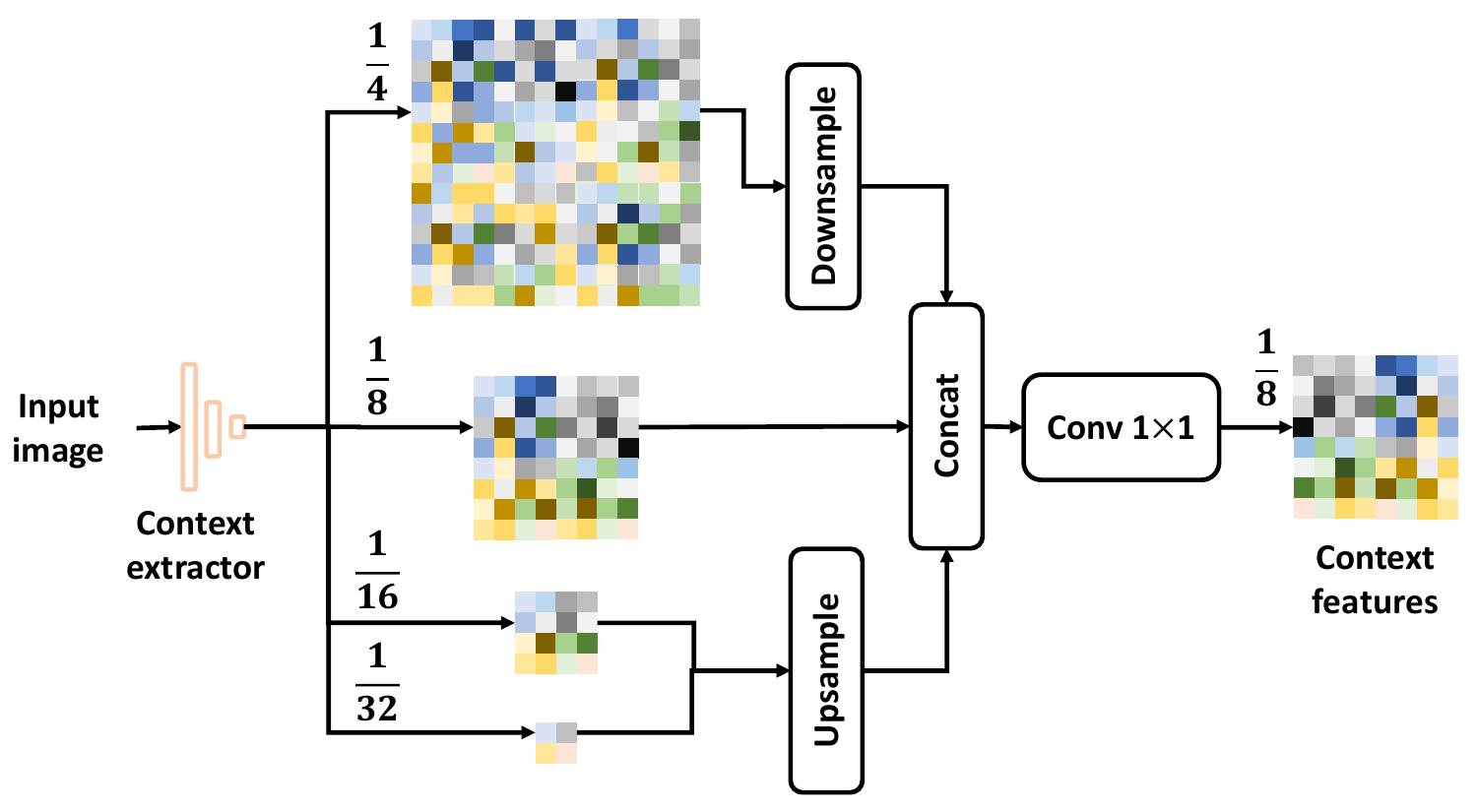}
    \caption{Overview of context extractor.}
    \label{fig:context_extractor}
\end{figure}

The difference between the transformation $\trans$ prediction process of our motion network and RAFT3D is summarized in this paragraph. As shown in \cref{eqn:gauss_netwon_step}, the similarities between rigid embeddings of pixels critically determine if two pixels are grouped as the same object. Thus, generating high-quality rigid embeddings has a direct impact on motion accuracy. Originally, RAFT3D uses a pre-trained ResNet-50 \cite{he2016deep} as the context extractor to provide semantics information. Instead, we replace the pre-trained ResNet-50 with a network similar to HRNet \cite{wang2020deep} that aggregates the context information across hierarchical levels. An overview is shown in \autoref{fig:context_extractor}. The context extractor extracts the semantic information from four resolutions -- $1/4, 1/8, 1/16$, and $1/32$ resolutions. We use bilinear upsampling to resize all feature maps to the $1/8$ resolution and use one $1\times1$ convolution to aggregate the information. Compared to ResNet-50, which has 40M parameters, our context extractor only has 3M parameters. We found that our motion network outperforms RAFT3D on the FlyingThings3D \cite{mayer2016large} dataset while having 1/5-th of the parameters of RAFT3D. In our implementation, we use the RGB image as semantic input instead of features from stereo to reduce dependency on the stereo network.

Once inter-frame motion $\trans$ is recovered, we use differentiable rendering \cite{ravi2020pytorch3d} to forward-project all feature maps \cite{li2020robotic} from the previous frame to the current frame, which is equivalent to Lagrangian flow where pixels of the past frame are \textit{pushed} to the current frame. This is in contrast to Eularian flow, where pixels are \textit{pulled} \cite{badin2018variational}, which is proposed in Spatial Transformer \cite{jaderberg2015spatial} and implemented as the ``grid\_sample'' function in PyTorch \cite{paszke2019pytorch}.

We evaluate the accuracy of the transformation map $\trans$ of our motion network because accurate motion prediction is critical to our process. We compare the performance of our motion network and RAFT3D in \cref{tab:motion}. To quantify the accuracy of $\trans$, we report the flow EPE ($\fepenull$) on optical flow $\fepe{of}$ and scene flow $\fepe{sf}$, and threshold metrics of 1px following \cite{teed2021raft}. We evaluate the performance in two settings, 1) taking the ground truth disparity as input and 2) taking the noisy stereo disparity estimates as input. For a fair comparison, we re-train RAFT3D using our training setup and report its result. \ours{} performs better than RAFT3D in both $\fepe{of}$ and $\fepe{sf}$ regardless of the input type, with only 1/5-th of the parameters of RAFT3D.

\setlength\tabcolsep{.5em}
\begin{table}[b]
\centering
\caption{Ablation study of the transformation prediction $\trans$ of the motion network. GT: ground truth disparity as input. S: stereo network disparity estimates as input. $\ddagger$: official checkpoint.}
\label{tab:motion}
\resizebox{\linewidth}{!}{%
\begin{tabular}{c|cc|cc|c}
\toprule
 & $\fepe{of} $ & $\dmetricsup{1px}{of} $ & $\fepe{sf} $ & $\dmetricsup{1px}{sf} $ & Parameters \\ \midrule
RAFT3D (GT) \cite{teed2021raft} $\ddagger$ & 2.145 & 0.131 & 2.177 & 0.138 & 45.0M \\ 
RAFT3D (GT) \cite{teed2021raft} & 1.808 & 0.133 & 1.847 & 0.141 & 45.0M \\
\bf Motion (Ours, GT) & \bf 1.754 & \bf 0.127 & \bf 1.793 & \bf 0.135 & \bf 8.5M \\ \midrule
RAFT3D (S) \cite{teed2021raft} & 2.458 & 0.149 & 2.514 & 0.158 & 45.0M \\
\bf Motion (Ours, S) & \bf 1.902 & \bf 0.134 & \bf 1.949 & \bf 0.142 & \bf 8.5M
\end{tabular}%
}
\end{table}

\subsection{Fusion Network}
\label{app:mem_network}
The fusion network uses a set of input cues to determine the $\mathbf{w}_\text{reset}$ and $\mathbf{w}_\text{fusion}$. 
For appearance correlation, we project the input features from the stereo network to a feature dimension of 32. The fusion weight is then regressed at $1/4$ resolution for better context awareness while the reset weight is regressed at the full resolution for better outlier rejection.

\setlength\tabcolsep{.5em}
\begin{table}[bt]
\centering
\caption{Results with different stereo networks on FlyingThings3D \cite{mayer2016large}.}
\label{tab:additional_result}
\resizebox{\linewidth}{!}{%
\begin{tabular}{c|cc|cc|cc}
\toprule
 & $\tepe$ & $\dmetricsup{3px}{t}$ & $\teper$ & $\dmetricsup{100\%}{t}$ & EPE & $\dmetric{3px}$  \\ \midrule
STTR & 0.482 & \bf 0.014 & 11.434 & 0.374 & \bf 0.449 & \bf 0.014 \\
\bf STTR + \ours{} & \textbf{0.448} & 0.017 & \textbf{10.109} & \textbf{0.290} & 0.454 & \bf 0.014 \\ \midrule
PSMNet & 1.371 & 0.056 & 35.136 & 0.466 & 1.079 & 0.045 \\ 
\bf PSMNet + \ours{} & \textbf{1.266} & \textbf{0.052} & \textbf{31.958} & \textbf{0.354} & \textbf{1.052} & \textbf{0.044} \\ \midrule
GwcNet & 0.959 & 0.041 & 22.598 & \textbf{0.409} & 0.752 & \textbf{0.032} \\
\bf GwcNet + \ours{} & \textbf{0.924} & \textbf{0.039} & \textbf{19.953} & 0.402 & \textbf{0.726} & \textbf{0.032} \\ 
\end{tabular}%
}
\end{table}

\section{Qualitative Visualization}
\label{app:qualitative_vis}
\subsection{Space-time volume}
\begin{figure}[b]
    \centering
    \includegraphics[width=\linewidth]{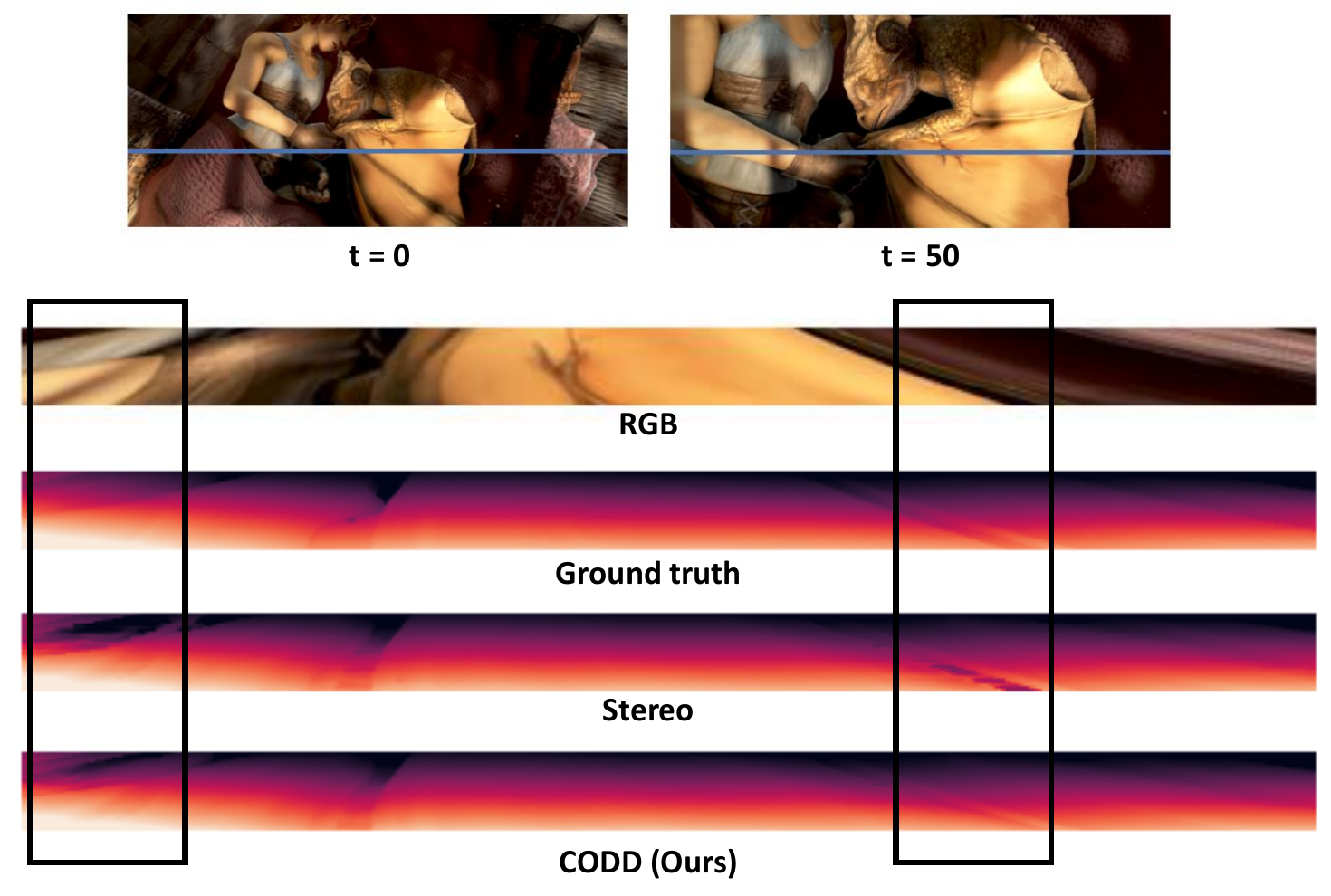}
    \caption{Space-time volumes created from MPI Sintel dataset \cite{butler2012naturalistic}. (a)-(b) RGB images of the start and end of the video sequence. The blue line indicates the horizontal slice taken. (c) RGB of the space-time volume slice. (d)-(f) Disparity of the space-time volume slice. The high-frequency noise of the stereo prediction is successfully removed with our CODD framework, as highlighted in the black boxes. \\ \textcopyright \space copyright Blender Foundation $|$ durian.blender.org}
    \label{fig:xtd_slice}
\end{figure}

Similar to \cite{zhang2021consistent}, we visualize the network's output by concatenating the predictions over the temporal axis to build a space-time volume on the MPI Sintel dataset. We then take a slice along the horizontal axis and visualize the depth prediction over time in \cref{fig:xtd_slice}. Even though this does not trace the same object point over time, it gives insights into how stable the model's prediction is temporally. We find that \ours{} contains less high-frequency variation than the per-frame model. 

\subsection{Improvements over Per-frame Stereo}
\begin{figure}[b]
    \centering
    \includegraphics[width=0.65\linewidth]{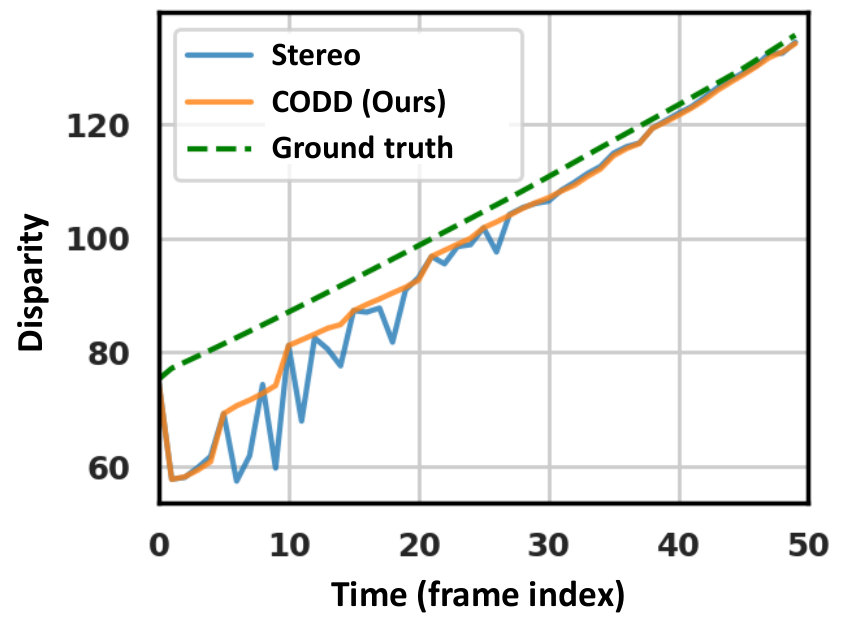}
    \caption{Disparity when tracing the same 3D point across time.}
    \label{fig:qual_vis}
\end{figure}

We visualize the improvements over stereo qualitatively in \autoref{fig:qual_vis}. As ground truth disparity constantly increases, the stereo predictions vary frequently. By incorporating temporal information, \ours{} removes many of the temporal jitters of the stereo prediction and outputs more consistent and accurate estimates. 

\begin{figure}[b]
    \centering
    \includegraphics[width=\linewidth]{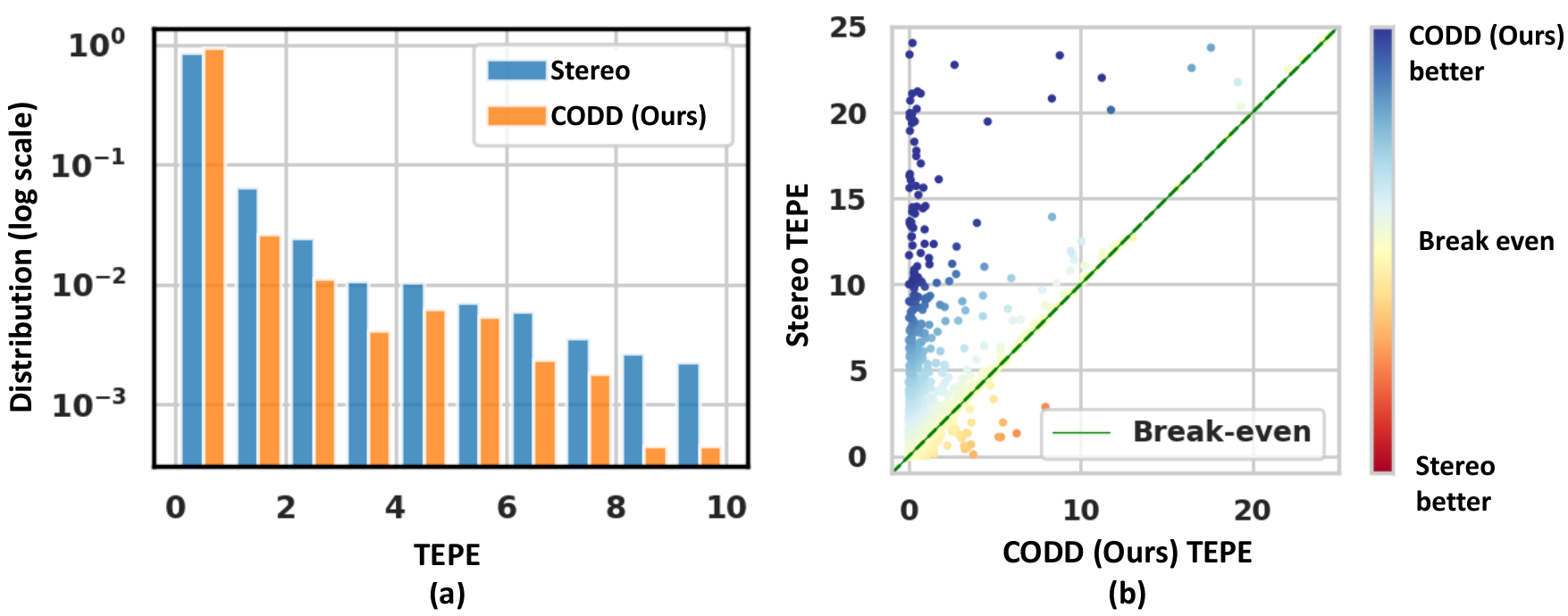}
    \caption{(a) Distribution of $\tepe$. (b) Pixel-wise improvement comparison. The diagonal line indicates break-even.}
   \label{fig:comparison_tepe}
\end{figure}

We plot the distribution of $\tepe$ comparing our and stereo's results in \autoref{fig:comparison_tepe}. We find that \ours{} successfully reduce $\tepe$ across different magnitudes as shown in \autoref{fig:comparison_tepe}a, suggesting more consistent predictions across time. To understand how the performance of each pixel has changed instead of the whole image, we additionally plot the $\tepe$ of each pixel of the stereo network and \ours{} in \cref{fig:comparison_tepe}b with one-to-one correspondence, where the diagonal line indicates break-even (\ie, $\tepe$ of both networks are the same), top left region indicates that \ours{} successfully reduces the $\tepe$, and bottom right region indicates that \ours{} makes the $\tepe$ larger. As most of the plotted points are clustered in the top left region, \ie, our $\tepe$ is smaller than stereo network $\tepe$, \ours{} reduces $\tepe$ of the stereo network for the majority of the pixels. We note that even in the cases where the $\tepe$ of the stereo network is large (y-axis),  \ours{} can reduce the error close to zero (x-axis), suggesting large improvements in terms of temporal consistency over the stereo network for these cases.

\section{Additional Experiments}
\label{app:additional_experiments}
We include additional experiments that shed light on \ours{} in the following sections.

\subsection{Using Different Stereo Networks}
\label{app:other-stereo}
We present additional experiments on FlyingThings3D \cite{mayer2016large} to demonstrate that our framework works with different stereo networks. We follow the same experimental setup as Sect.5.2 and use other stereo networks to produce the per-frame disparity estimates. \cref{tab:additional_result} summarizes our findings, which are consistent with our findings for HITNet. We note that all metrics improve with the introduction of our motion and fusion networks, with the only exceptions of STTR $\dmetricsup{t}{3px}$, EPE and GwcNet $\dmetricsup{100\%}{t}$. The results suggest that our proposed framework should apply to other stereo networks.

\subsection{End-to-End Fine-tuning}
\label{app:e2e_training}
We note that our framework is designed to be end-to-end differentiable. Thus, during fine-tuning, we can in theory directly fine-tune all components together, instead of first fine-tuning the stereo network and then the other components. However, due to the gradients from different losses back-propagating to the stereo network, in comparison to only the stereo losses in the case of a per-frame stereo network, end-to-end fine-tuning may not be a fair comparison. Thus, we report these additional results in the appendix. 

Denoting the loss of stereo network as $\xstereo{\ell}$, the loss of motion network as $\xmotion{\ell}$, and the loss of fusion network as $\xfusion{\ell}$, during end-to-end fine-tuning, the total loss is the weighted sum of all sub-network losses:
\begin{equation}
    \ell_{\text{E2E}} = \xstereo{\alpha}\xstereo{\ell} + \xmotion{\alpha} \xmotion{\ell} + \xfusion{\alpha}\xfusion{\ell}\,.
    \label{eqn:loss_e2e}
\end{equation}
We use $\xmotion{\alpha}=0.5$ while keeping $\xstereo{\alpha}= \xfusion{\alpha}=1.0$ in our experiments to balance the losses. A batch size of 16 is used for end-to-end fine-tuning due to memory constraints. All learning rates are linearly decayed from $2\mathrm{e}{-5}$ following the learning rate of the per-frame stereo model.

The results on different benchmarks are summarized in \cref{tab:e2e}, where the end-to-end training (E2E) result is comparable to the stage-wise fine-tuning results in Sect.5.3. Other than TartanAir, end-to-end fine-tuning leads to better results. In either training setup, our results are favorable compared to the per-frame results. The results suggest that our framework is not sensitive to the specific training strategies we have used.

\setlength\tabcolsep{.5em}
\begin{table}[bt]
\centering
\caption{End-to-end fine-tuning results. Ours: Fine-tune the stereo network first and then freeze its parameters. Ours (E2E): end-to-end fine-tuning.}
\label{tab:e2e}
\resizebox{\linewidth}{!}{%
\begin{tabular}{c|c|cc|cc|cc}
\toprule
  \multicolumn{2}{c|}{} & $\tepe $ & $\dmetricsup{3px}{t} $ & $\teper $ & $\dmetricsup{100\%}{t} $ & EPE & $\dmetric{3px} $ \\ \midrule
    \multirow{3}{*}{\shortstack{Tartan Air \\ dataset \cite{wang2020tartanair}}} & Stereo \cite{tankovich2021hitnet} & 0.876 & 0.053 & 9.039 & 0.339 & 0.934 & 0.055  \\ 
    & \bf \ours{} & \bf 0.751 & \bf 0.045 & \bf 6.206 & \bf 0.240 & \bf 0.904 & 0.053 \\
    & \bf \ours{} (E2E) & 0.763 & 0.047 & 6.976 & 0.270 & 0.905 & \bf 0.052 \\ \midrule
   \multirow{3}{*}{\shortstack{KITTI Depth \\ dataset \cite{uhrig2017sparsity}}} & Stereo \cite{tankovich2021hitnet} & 0.289 & \bf 0.001 & 3.630 & 0.156 & 0.423 & 0.004  \\ 
   & \bf \ours{} & 0.258 & \bf 0.001 & 2.764 & 0.132  & 0.418 & \bf 0.003 \\ 
   & \bf \ours{} (E2E) & \bf 0.251 & \bf 0.001 & \bf 2.408 & \bf 0.129 & \bf 0.409 & \bf 0.003 \\ \midrule
   \multirow{3}{*}{\shortstack{KITTI 2015 \\ dataset \cite{menze2015joint}}} & Stereo \cite{tankovich2021hitnet} & 0.570 & 0.026 & 10.672 & 0.126 & \multirow{2}{*}{0.811} & \multirow{2}{*}{0.033} \\ 
   & \bf  \ours{} & 0.507 & \bf 0.022 & 8.740 & 0.112 & &   \\ 
   & \bf \ours{} (E2E) & \bf 0.505 & 0.024 & \bf 8.132 & \bf 0.111 & \bf 0.806 & \bf 0.032
\end{tabular}%
}
\end{table}

\subsection{Zero-shot Generalization}
\label{app:zero-shot}
\textbf{Dataset} 
We report zero-shot generalization results of our pre-trained network from FlyingThings3D on the TartanAir, KITTI Depth, and MPI Sintel dataset finalpass \cite{butler2012naturalistic}. MPI Sintel contains animated characters with deformation. It contains 23 video sequences, totaling 1064 images. 

\textbf{Results} As shown in \cref{tab:zero-shot}, our model can generalize well onto a new data domain and significantly improves over the per-frame stereo \cite{tankovich2021hitnet} and Kalman filter \cite{welch1995introduction} in terms of temporal metrics, up to 36\% ($\teper$: from 298.674 to 191.445). \ours{} also outperforms the competing approaches in terms of EPE.

\setlength\tabcolsep{.5em}
\begin{table}[bt]
\centering
\caption{Zero-shot generalization experiments on MPI Sintel \cite{butler2012naturalistic} datasets. All networks are trained only on FlyingThings3D \cite{mayer2016large}.}
\label{tab:zero-shot}
\resizebox{\linewidth}{!}{%
\begin{tabular}{c|cc|cc|cc}
    \toprule
 & $\tepe$ & $\dmetricsup{3px}{t}$ & $\teper$ & $\dmetricsup{100\%}{t}$ & EPE & $\dmetric{3px} $ \\ \midrule
Stereo \cite{tankovich2021hitnet} & 2.621 & 0.127 & 298.674 & 0.515 & 5.028 & 0.199 \\ 
Kalman Filter \cite{welch1995introduction} & 2.583 & 0.126 & 287.456 & 0.460 & 5.027 & 0.199 \\
\bf \ours{} (Ours) & \bf 2.270 & \bf 0.092 & \bf 191.445 &\bf  0.439 &\bf  5.009 & 0.199 \\
\end{tabular}%
}
\end{table}

\subsection{Additional Fusion Network Ablation}
\label{app:additional_fusion_ablation}
We provide additional ablation experiments to further understand how different design choices may affect the performance in \cref{tab:additional-ablation-memory}.

\vspace{1ex} \noindent \textbf{Correlation window} We experiment with different correlation window size $W$ to see the effect of increasing receptive field. We increase the correlation window size of correlation from 3 to 5 or 7, with a constant dilation of 2. We do not observe significant benefits of increasing the window size. The result suggests that a window size of 3 with dilation of 2 already provides enough information for the network. Our final model uses a window size of 3 for efficiency.

\vspace{1ex} \noindent \textbf{Correlation type} In addition to our proposed pixel-to-patch correlation, we also experiment with: 1) pixel-to-pixel correlation, where correlation values are obtained pixel-wise, and 2) patch-to-patch correlation, where correlation values are computed pixel-wise over the whole patch. We find that pixel-to-patch correlation performs best across all metrics.

\setlength\tabcolsep{.5em}
\begin{table}[bt]
\centering
\caption{Additional ablation studies of fusion network. \underline{Underline}: the base configuration.}
\label{tab:additional-ablation-memory}
\resizebox{\linewidth}{!}{%
\begin{tabular}{c|c|cc|cc|cc}
\toprule
\multicolumn{2}{c|}{} & $\tepe $ & $\dmetricsup{3px}{t} $ & $\teper $ & $\dmetricsup{100\%}{t} $ & EPE & $\dmetric{3px} $  \\ \midrule
Correlation & \underline{3} & 0.756 & 0.035 & 15.013 & 0.211 & 0.604 & 0.029 \\ 
window & 5 & 0.756 & 0.035 & 15.005 & 0.210  & 0.603 & 0.029 \\ 
$W$ & 7 & 0.756 & 0.035 & 15.014 & 0.210 & 0.603 & 0.029 \\ \midrule
\multirow{3}{*}{\shortstack{Correlation \\ Type}} & \underline{pixel-to-patch} & 0.756 & 0.035 & 15.013 & 0.212 & 0.604 & 0.029 \\ 
& pixel-to-pixel & 0.758 & 0.036 & 15.103 & 0.211  & 0.610 & 0.030 \\ 
& patch-to-patch & 0.759 & 0.035 & 15.134 & 0.210 & 0.609 & 0.030
\end{tabular}%
}
\end{table}

\section{Dataset Details}
\label{app:weight_sup_kitti}
For long sequences (several thousand), we chunk the video into sub-sequences of 50 frames following \cite{butler2012naturalistic}. 

\vspace{1ex} \noindent \textbf{FlyingThings3D}: We follow the official split information.

\vspace{1ex} \noindent \textbf{TartanAir}: We validate on the \textit{seasonforest} sequence and test on the \textit{carwelding} sequence.

\vspace{1ex} \noindent \textbf{KITTI Depth} \cite{uhrig2017sparsity}\footnote{http://www.cvlibs.net/datasets/kitti/eval\_depth\_all.php}: We follow the official split on the KITTI Depth dataset, except the \textit{People} scene for training due to little variation. We use the last training sequence \textit{2011\_10\_03\_drive\_0042\_sync} for validation. Due to the missing ground truth optical flow information, we use an off-the-shelf network to generate the optical flow information \cite{teed2020raft} and use forward-backward flow consistency to determine flow occluded regions and outliers. The disparity change is then computed from optical flow.

\vspace{1ex} \noindent \textbf{KITTI 2015} \cite{menze2015joint}\footnote{http://www.cvlibs.net/datasets/kitti/eval\_scene\_flow.php}: We use a 5-fold split that splits the data into train, validation, test data.

\section{Training Stereo Depth Network in Multi-frame Setting}
One of the simplest ideas we have tried to improve the temporal consistency of stereo depth networks is to train stereo networks in a multi-frame setting instead of randomly sampling images from the whole dataset. For a fair comparison, we reduce the cropping height by two times and insert a subsequent frame from the same video sequence, such that the total number of pixels that the network sees is the same. However, we find this does not perform well compared to the original training scheme as shown in \cref{tab:mf_training}, which may be attributed to the reduced diversity in the sample size.

\setlength\tabcolsep{0.5em}
\begin{table}[bt]
\centering
\caption{Comparison of training schemes for stereo depth network on the FlyingThings3D dataset \cite{mayer2016large}.}
\label{tab:mf_training}
\resizebox{0.7\linewidth}{!}{%
\begin{tabular}{c|c|c|c}
 & $\tepe$ & TEPE$_\text{r}$ & EPE \\ \midrule
 default training & \bf 0.821 & \bf 16.450 & \bf 0.609 \\ 
 multi-frame training & 0.998 & 19.199 & 0.715
\end{tabular}%
}
\end{table}

\section{Optical Flow as Alignment Mechanism}
One alternative to our motion network is to predict an inter-frame optical flow (\ie without disparity change compared to scene flow) to provide past information. However, since depth is not translation invariant, the past information can only provide past trending information instead of being directly aggregated. As a proof-of-concept, we replace the motion network with the ground truth optical flow provided to test the optical flow idea. However, we find that the temporal stability often doesn't benefit from the past information, and in fact, becomes worse on the FlyingThings3D dataset. 

\end{document}